\documentclass[letterpaper, 10 pt, conference]{ieeeconf}  

\IEEEoverridecommandlockouts                              

\usepackage{graphics} 
\usepackage[dvips]{graphicx}

\usepackage[caption=false,font=footnotesize,labelfont=sf,textfont=sf]{subfig}
\usepackage{color}

\usepackage{url}
\usepackage{tabularx, booktabs}
\newcolumntype{Y}{>{\centering\arraybackslash}X}

\usepackage{amsmath}
\usepackage{color,soul}

\usepackage{flushend}
\usepackage{tikz}

\newcommand\copyrighttext{%
	\footnotesize \textcopyright 2022 IEEE. Personal use of this material is permitted.
	Permission from IEEE must be obtained for all other uses, in any current or future
	media, including reprinting/republishing this material for advertising or promotional
	purposes, creating new collective works, for resale or redistribution to servers or
	lists, or reuse of any copyrighted component of this work in other works.
	DOI: \href{https://doi.org/10.1109/ICRA46639.2022.9812245}{10.1109/ICRA46639.2022.9812245}}
\newcommand\copyrightnotice{%
	\begin{tikzpicture}[remember picture,overlay]
		\node[anchor=south,yshift=10pt] at (current page.south) {\fbox{\parbox{\dimexpr\textwidth-\fboxsep-\fboxrule\relax}{\copyrighttext}}};
	\end{tikzpicture}%
}

\title{\LARGE \bf
Neural Style Transfer with Twin-Delayed DDPG\\* for Shared Control of Robotic Manipulators\vspace*{-2mm}
}

\author{  Raul Fernandez-Fernandez$^{1*}$, Marco Aggravi$^{2}$, Paolo Robuffo Giordano$^{2}$,\\ Juan G. Victores$^{1}$  and Claudio Pacchierotti$^{2}$ 
\thanks{$^{1}$Robotics Lab, Department of Systems Engineering and Automation,
    Universidad Carlos III de Madrid (UC3M), e-mail: \{rauferna, jcgvicto\}@ing.uc3m.es.}%
\thanks{$^{2}$CNRS, Univ Rennes, Inria, IRISA -- Rennes, France, e-mail: \{marco.aggravi, prg, claudio.pacchierotti\}@irisa.fr.}%
\thanks{$^{*}$ Corresponding author: rauferna@ing.uc3m.es.}%
}

\makeatletter
\let\NAT@parse\undefined
\makeatother

\usepackage[pdftex, plainpages=false,hypertexnames=true,pdfnewwindow=true,backref=true,colorlinks=true,citecolor=blue,linkcolor=red,urlcolor=blue,filecolor=blue]{hyperref}%

\begin{document}
\maketitle
\copyrightnotice

\begin{abstract}

Neural Style Transfer (NST) refers to a class of algorithms able to manipulate an element, most often images, to adopt the appearance or style of another one. Each element is defined as a combination of Content and Style: the Content can be conceptually defined as the ``what'' and the Style as the ``how'' of said element. In this context, we propose a custom NST framework for transferring a set of styles to the motion of a robotic manipulator, e.g., the same robotic task can be carried out in an ``angry'', ``happy'', ``calm'', or ``sad'' way. An autoencoder architecture extracts and defines the Content and the Style of the target robot motions. A Twin Delayed Deep Deterministic Policy Gradient (TD3) network generates the robot control policy using the loss defined by the autoencoder. The proposed Neural Policy Style Transfer TD3 (NPST3\footnote[3]{NPST3: Neural Policy Style Transfer TD3}) alters the robot motion by introducing the trained style. Such an approach can be implemented either offline, for carrying out autonomous robot motions in dynamic environments, or online, for adapting at runtime the style of a teleoperated robot. The considered styles can be learned online from human demonstrations. We carried out an evaluation with human subjects enrolling 73 volunteers, asking them to recognize the style behind some representative robotic motions. Results show a good recognition rate, proving that it is possible to convey different styles to a robot using this approach.

\end{abstract}


\section{Introduction}

Neural Style Transfer (NST) was proposed by Gatys \emph{et al.} \cite{Gatys2016} to define the Content and Style of an image using neural networks. Gatys \emph{et al.} used features extracted by a VGG-19 pre-trained neural network~\cite{Simonyan2014} to define the Content as the elements represented in the painting (e.g., people, animals, houses), and the Style as the low-level features specific to the artist (e.g., vivid colors, long brushstrokes). 

One advantage of NST is the definition of a high-level layer of abstraction that allows the same definition of Content and Style in a larger range of applications. One example is the area of motion animation \cite{Holden2017}. Here the Content can be defined as the movements performed by the animated figure (e.g., moving forward, walk in circles, move sideways) and the Style as the emotion conveyed by these movements (e.g., sad, happy, tired, angry). Holden \emph{et al.} \cite{Holden2017} proposed the introduction of autoencoders for working with motions. The result is a NST generated motion trajectory that can be transferred to the animated figure. To the best of our knowledge, such an approach has never been applied to robotic autonomous control and teleoperation.

\begin{figure}[t]
  \vspace{1.5mm}
  \centering
  \includegraphics[width=0.39\textwidth]{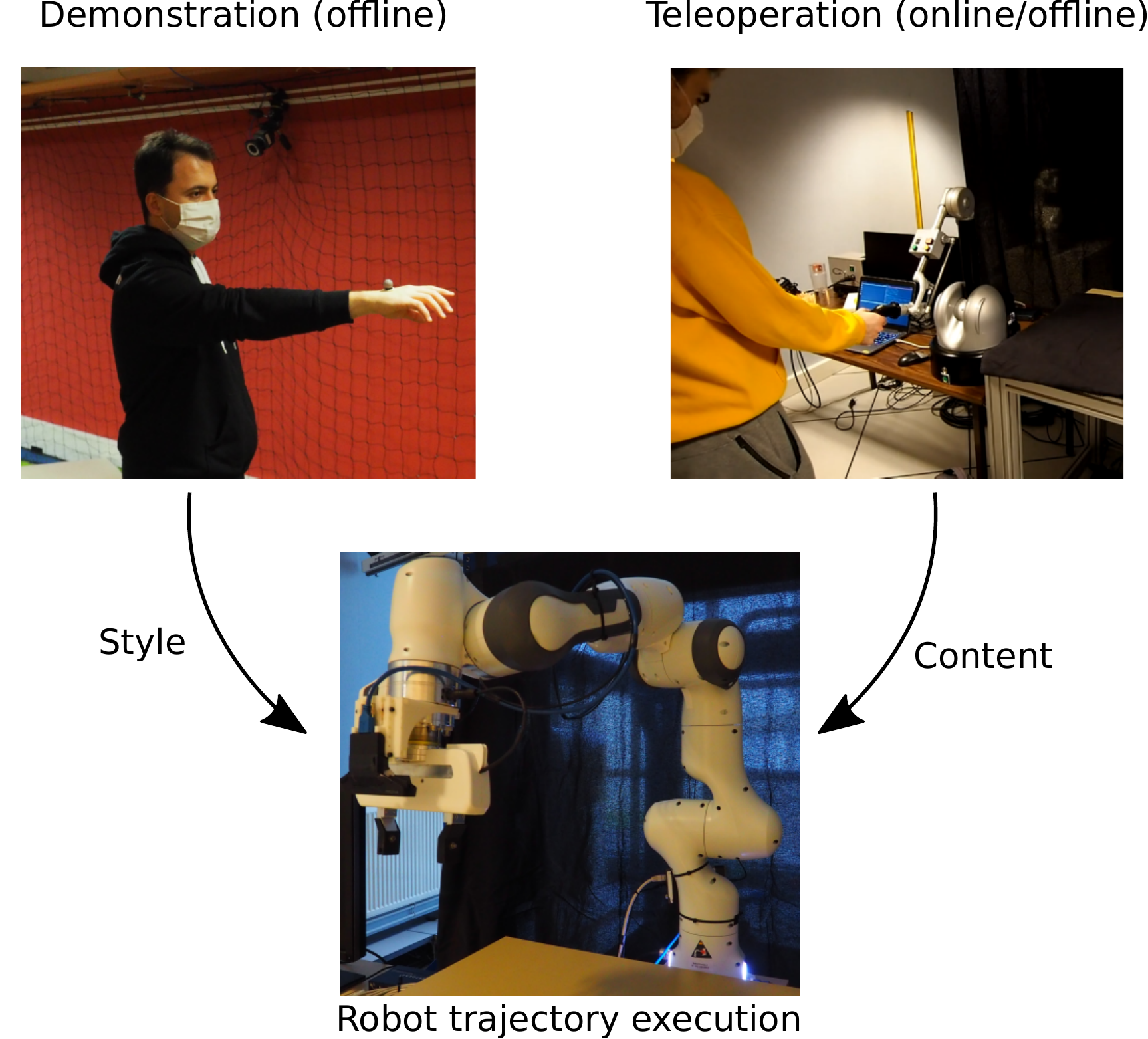} 
  \vspace*{-3mm}
  \caption{Using Neural Style Transfer (NST), we can alter a robotic motion (the Content) according to the Style of a pre-recorded human demonstration. A certain teleoperated robotic motion can be carried out in, e.g., a angry, happy, calm, or sad way. }
  \label{fig:idea}
  \vspace*{-6mm}
\end{figure}

Previous works of the authors \cite{FernandezFernandez2022} have proposed an initial approach for introducing NST within discrete action spaces using Q-Networks. In this paper, we propose the use of a Neural Policy Style Transfer Twin Delayed Deep Deterministic Policy Gradient (NPST3) framework for introducing NST
in continuos robotic action spaces, with teleoperation, and autoencoders to extract the Style Transfer loss. DRL has already been used to generate robust robotic controllers \cite{Levine2016}, while Deep Deterministic Policy Gradient (DDPG) \cite{lillicrap2016} is an application of DRL within continuous action spaces. Finally, Twin Delayed DDPG (TD3) \cite{Fujimoto2018} has shown promising results as an advanced and more reliable approach to DDPG. 

Our objective is to abstract the Style of a human motion and then apply it to the Content of an unrelated robotic motion. The latter can be generated offline by an autonomous controller or online by a human operator teleoperating the target robot. Autoencoders are introduced to define the Content and Style for training, while the TD3 algorithm is used to generate the robot control policies and encode the Style Transfer optimization step. The Content is defined as the high-level features that define the robot action (e.g., end-point of the trajectory) while the Style is defined as the low-level features specific to a certain human demonstration (e.g., speed, jerkiness). As a proof of concept, we consider four different Styles defined via human demonstrations: angry, happy, calm, and sad motions. Similarly, we define the Content as the control motion imparted by a human operator in teleoperation. Doing so, a human user teleoperates a robot that, in turn, actuates the overall imparted motion in a angry, happy, calm, or sad way, locally deviating from the teleoperated Content motion to apply the chosen Style. The framework idea is depicted in Fig. \ref{fig:idea}.

Being able to impart a certain Style to a robotic motion via NST opens a new area of possible applications and can be of interest for art performances, animatronics, robot caregivers and waiters in smart city applications \cite{fernandez-fernandez2018}, craftsmanship, and in any other situation where a personalized motion is somehow important. An implementation of our approach, including all the statistics parameters we considered, is available at \protect\url{https://github.com/RaulFdzbis/NPST3}.

\section{Background and Preliminaries}
\label{Section:Background}

Style Transfer is used to extract the Content and Style of the motions, while DRL permits to execute and generate the control policies defining the new robotic motions. 

\subsection{Style Transfer}
\label{Section:Back_ST}

The introduction and modification of Style in motions is not a new topic within the computer animation community~\cite{Bruderlin1995}. The first works introducing Style in the area of animation proposed signal processing techniques for the definition and extraction of the Style \cite{Unuma1995}. More recent works proposed the implementation of more advanced techniques as multilinear model design \cite{Min2010}. The first works to differentiate between Content and Style, however, were part of the area of optical character recognition \cite{tenenbaum1997}. In robotics, Style Transfer has been related to the introduction of emotions in robotic motions. These works include the introduction of cost functions \cite{Zhou2018} or Laban movement systems \cite{Sharma2013}. 

With the publication of NST by Gatys \emph{et al.} \cite{Gatys2016}, a new layer of abstraction was introduced through the introduction of the VGG-19 pre-trained classification neural network. Due to the difficulty of having a proper motion classification pre-trained neural network, Holden \emph{et al.} \cite{Holden2017} proposed the introduction of autoencoders as an alternative. 
Autoencoders are neural networks that can be divided in two parts: the encoder and the decoder. The encoder is composed by the first layers of the network in charge of extracting the relevant features to generate the encoded data. An encoder operation can be defined following $A(X) = ReLU(\Psi(X*W_0+b_0))$,
where $X$ is the input, $\Psi$ is the pooling operation posterior to the first layer, $W_0$ is the weight matrix of the encoder, $b_0$ is the layer bias, and $ReLU$ is a Rectified Linear Units (ReLU) \cite{Nair2010} activation. The decoder, corresponding to the last layers, takes the encoded data and performs the encoder inverse operation to regenerate the input. 
The Content loss can then be defined as the difference in the encoder outputs when passing the content and generated motions following 

\begin{equation}\label{eq:Lc} 
L_{content}=\|A(C) - A(G)\|, 
\end{equation}

where $C$ and $G$ are the Content and Generated motions, respectively. The Style loss is defined as the difference between the Gram matrix $Gm$ of the encoder outputs when passing the style motion and Generated motions as in 

\begin{equation}\label{eq:Ls} 
L_{style}=\|Gm(A(S)) - Gm(A(G))\|, 
\end{equation}

where $S$ is the Style motion. Finally, the total Style Transfer loss is defined as the sum of these two losses $L_{st} =L_{content}+L_{style}$.

\subsection{Deep Reinforcement Learning}
 
Robotics tasks usually require continuous and high-dimensional action spaces, while standard Reinforcement Learning techniques, like Q-learning \cite{Watkins1992}, can only be directly applied to discrete action spaces.  To address this limitation, Lillicrap \emph{et al.} \cite{lillicrap2016} proposed the introduction of DDPG for continuous action spaces. In this approach, the policy is defined as a parametric function. This parametric function is trained to maximize the output of the Q-value function. The authors proposed the introduction of two different neural networks within an actor-critic architecture to define the policy and the Q-value function. Based on this idea, TD3 was later proposed by Fujimoto \emph{et al.} \cite{Fujimoto2018} as a more stable version of DDPG. TD3 improves the performance of DDPG by introducing Double Q-learning \cite{Hasselt2015}, target policy smoothing, and delayed policy updates.

\begin{figure*}[h!t]
	\vspace*{2mm}
	\centering
	\includegraphics[width=0.9\textwidth]{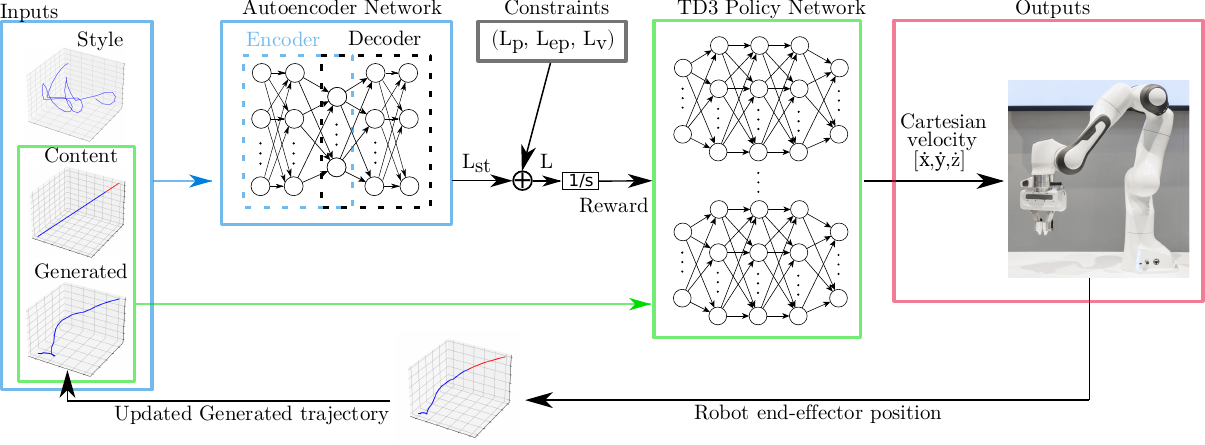} 
	\vspace*{-3mm} 
	\caption{Framework for the NPST3 algorithm. The Style, Content, and Generated motion trajectories are used as input for the autoencoder network. The loss obtained with the autoencoder network ($L_{st}$) is added to the resulting constraint loss ($L_{p}$, $L_{ep}$, $L_{v}$) as to obtain the overall loss $L$. The inverse (1/s) of this loss is the reward used for training the TD3 Policy network, together with  the Content and Generated trajectories as input. The result of the TD3 network is a 3D Cartesian velocity vector, which is executed by the robot end effector. Finally, the Generated trajectory is updated adding the new position actuated by the robot, while the Content trajectory is updated using the positions defined by the user. The Content trajectory can be defined online via teleoperation or offline via a preplanned motion. }
	\vspace*{-3mm}
	\label{fig:graph}
\end{figure*}

\section{Framework}
\label{sec:framework}

The framework is depicted in Fig.~{\ref{fig:graph}}. The Style Transfer block with the autoencoder network (blue) defines the Style Transfer loss, while the execution block with the TD3 policy network (green) minimizes the obtained Style Transfer and constraints losses. After training, the resulting TD3 policy network encodes the Style Transfer step, and it is the only network required for the control of the robot (red).

\subsection{Inputs}

The inputs of the framework are three 3D Cartesian motion trajectories that define the Content, the Style, and the Generated motions. 
The total length of all the input trajectories is fixed to 5 seconds. The number of samples per second is set to 10.
These settings are a trade-off between motion accuracy and computational cost, as the motions need to be Generated at runtime. The motion trajectory matrices ($C$,$S$,$G$) used as inputs, have a $[m,n]$ shape, where $m$ is the total number of samples, in this case 50, and $n$ is the space dimension, in this case 3. The framework is designed to be able to work with incomplete motions (trajectories defined with less than 50 samples) allowing the Generated and Content motions to be used as input while being generated.

\subsection{Autoencoder network: the loss network}
A convolutional autoencoder network defines the loss network. Following the architecture proposed by Holden \emph{et al.} \cite{Holden2017}, the encoder is defined using a 1D convolutional layer followed by a pooling operation layer. The convolutional layer is composed by 256 nodes and a kernel size of 5.  The two layers of the decoder are defined as the transpose layers of the convolutional and pooling layers of the encoder. For the encoder, an additional dropout layer is introduced to improve the performance. The resulting network is used to extract the content and style loss as in Eq. \ref{eq:Lc} and \ref{eq:Ls}.

\subsection{Constraints}
Similarly to Holden \emph{et al.} \cite{Holden2017}, in addition to the loss defined by the loss network, a few additional constraints are introduced to generate feasible and acceptable motions for the robot, e.g., the generated trajectory should be as similar as possible to the one commanded by the operator, there should no be discontinuities between trajectory executions, or the velocity of the generated trajectory should be bounded. 

The first constraint limits the position error of the Generated motion with respect to the Content motion.

\begin{equation}\label{eq:PL} 
L_{p} = \|\frac{G[t-1]-C[t-1]}{RT}\|,
\end{equation}
where $t$ is the current execution step and the Robot Threshold (RT) is a handcrafted constant to normalize the values of the Cartesian workspace, assumed the same for all axes.

The second constraint limits the position error introduced on the last step of the motion trajectory. This error is introduced to smooth the transition between consecutive motions. This constraint is introduced following
\begin{equation}\label{eq:EPL} 
L_{ep} = \|\frac{G[t_{n}]-C[t_{n}]}{RT}\|,
\end{equation}
where $t_{n}$ is the last time step. This loss is set to $0$ if the last time step has not been reached.

The third and last constraint is a velocity constraint introduced to increase the weight of the velocity for the Generated motion. This constraint is defined as

\begin{equation}\label{eq:VL} 
L_{v} = \|(\frac{\partial G}{\partial t}-\frac{\partial S}{\partial t})/RT\|.
\end{equation}

The total loss is defined as a weighted sum of these constraints and the Style Transfer losses as in

\begin{equation}\label{eq:TL} 
L = w_{c}L_{content}+w_{s}L_{style}+w_pL_{p}+w_{ep}L_{ep}+w_{v}L_{v}
\end{equation}

This loss will be used later for training the execution network.


\subsection{TD3 Policy network: the execution network} 

The execution network is a TD3 network in charge of the generation of the Generated motion that minimize the loss defined in the Style Transfer stage. The inputs of the network are the Content and Generated motion trajectories. Both trajectories are incrementally introduced to the network. This allows the online generation of the Content motion. 
We need to train a different network for each style; however, all of these style-tailored networks will work for any Content.

The TD3 architecture is composed by two different sub-networks: the actor and the critic \cite{Konda2000}. The actor network encodes the control policy of the robot, while the critic encodes the Q-value function. For the actor network, each input passes through three 1D convolutional layers before being concatenated with the other. The first convolutional layer is composed by 256 nodes and a kernel size of 5. This layer was designed to have the same configuration than the one used in the encoder. The rest of the convolutional layers have 128 nodes and a kernel size of 5. After these three layers, the two outputs of these convolutional layers are flattened and concatenated. The concatenated output is passed through four fully connected layers. The first two layers have a total of 512 nodes, the third 400, and the last one 300. A batch normalization layer is added between all the layers of the network. A ReLU activation is used for all the layers except the last one which uses an hyperbolic tangent (tanh) activation. The output of the network is a 3D velocity vector of the robot end-effector.

The critic network has the same architecture as the actor network with some exceptions. First, an additional input is introduced for the action, defined as the output of the actor network. This action is not passed through the convolutional layers, but instead directly through the fully connected layers. After the first fully connected layer, the action is concatenated with the two other inputs. An additional fully connected layer of size 512 is added before the 400 size layer. Finally, the last layer of the network uses a linear activation function and its output corresponds to the Q-value. The specific design choices reported have been chosen during pilot experiments.

\subsection{Outputs}

The output of the framework is a 3D Cartesian velocity vector. This velocity vector is used to command the end effector of the selected robotic platform, e.g., a robotic arm in our case. The new end effector position, obtained after the execution of the velocity vector, is used to update the Generated motion trajectory. This updated Generated motion trajectory is used as input for the next step of the framework.  
This approach can be used with any robotic platform that can be controlled through velocity commands.

\section {Training}

Similarly to the work proposed by Li \emph{et al.} \cite{Li2019}, to train the autoencoder we used the CMU Graphics Lab Motion Capture Database \cite{CMU2003}. This dataset contains various motions performed by different people, captured using a motion capture system. For simplicity, only the information from the tracker at the end of the right hand (\emph{RFIN}) was considered. 


\begin{figure}[h]
  \vspace{2mm}
  \centering
    \includegraphics[width=0.48\textwidth]{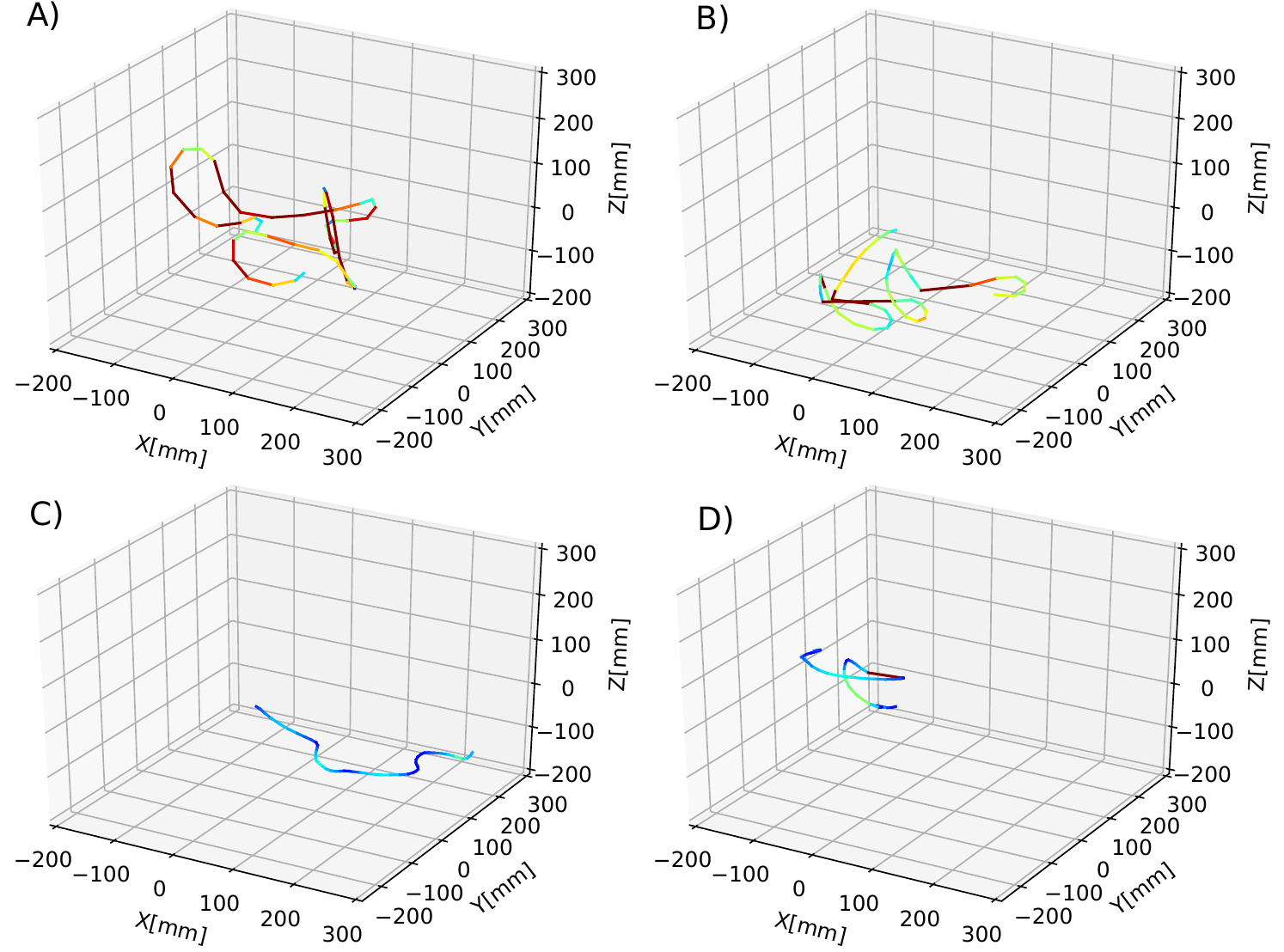}
    \includegraphics[width=0.48\textwidth]{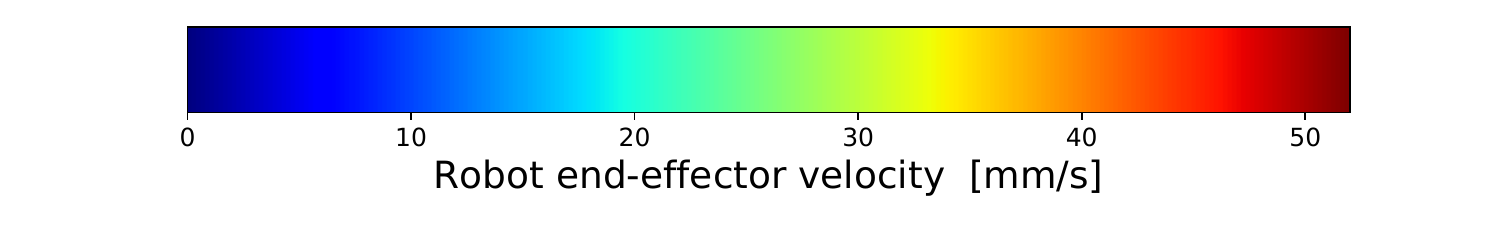}
  \vspace*{-6mm}
  \caption{Cartesian trajectories of the selected Styles. A) anger/annoyance, B) happiness/joy, C) calm/acceptance, and D) sadness/grief. The units in the axis are in mm. }
  \label{fig:all-styles}
  \vspace{-3mm}
\end{figure}

In the case of the execution network, a random generator of linear motions was implemented to generate the training Content motions. We considered four Styles, related to the emotions of anger/annoyance, happiness/joy, calm/acceptance, and sadness/grief, which are part of the wheel of basic emotions as defined by Plutchik~\cite{Plutchik1980}.
Such Styles were defined via human demonstrations:
a single person was asked to move his right hand freely in space, so as to convey -- in any ways he wanted to -- each of the four emotions mentioned above.
The position of the user's hand over time was recorded using a Vicon motion capture system.
For the training, a 5-seconds-long portion of the full demonstrated motion was selected for each Style. Four execution networks were then trained, each corresponding to one Style. Finally, a TD3 algorithm was implemented for the training. The loss described by Eq. \ref{eq:TL} was selected as the loss function to minimize. The training parameters of the two networks are listed in Table~\ref{tab:parameters}. They were defined after a set of preliminary experiments consisting of 50 tentative trainings, retaining the parameters best expressing the target emotions as identified by the user who carried out the demonstrations.

\begin{table}[t]
\vspace{0.5em}
\centering
\caption{Training parameters}
\label{tab:parameters}
\begin{tabular}{ll}
Hyperparameters                 & Values \\ \hline
\emph{Shared}                         &                                     \\
\hspace{3ex} Motion input shape             & [50,3]                             \\
\hspace{3ex} Sample frequency               & 10 [Hz]                           \\
\hspace{3ex} Motion length                  & 5 [s]                          \\
\hspace{3ex} Motion Robot Threshold (RT)    & 300 [mm]                          \\
\hspace{3ex} Optimizer                      & Adam \cite{Kingma2014}    \\
\hspace{3ex} Number of styles               & 4    \\
\emph{Autoencoder}                  &                                     \\
\hspace{3ex} Epochs                         & 1000                               \\
\hspace{3ex} Batch size                     & 256                              \\
\emph{TD3 network}                  &                                     \\
\hspace{3ex} Action space dimensions    & 3                         \\
\hspace{3ex} Action Range (AR)              & $\pm$ 0.1 * RT                 \\
\hspace{3ex} Loss weights $(w_c, w_s, w_p, w_{ep}, w_v) $             & (100, 1, 0.1, 1, 20)    \\
\hspace{3ex} Epochs                         & 2500                                \\
\hspace{3ex} Experience Replay size         & 10e3                                \\
\hspace{3ex} Batch size                     & 64                              \\
\hspace{3ex} Critic Learning Rate           & 1e-5                               \\
\hspace{3ex} Actor Learning Rate            & 1e-6                               \\
\hspace{3ex} Discount ($\gamma$)            & 0.99                               \\
\hspace{3ex} Critic/Actor update ratio      & 2                               \\
\hspace{3ex} Target update value (tau)      & 1e-3                               \\
\hspace{3ex} Loss Function                  & Mean Squared Error                  \\
\hspace{3ex} Initialization network values  & $\pm$ 3e-3 (Uniform)   \\
\hspace{3ex} Policy noise                   & 0.002 * RT (Normal)    \\
\hspace{3ex} Action noise                   & 0.02 * RT (Normal)    \\

\end{tabular}%
\vspace*{-4mm}
\end{table}

\section{Experiments}

A human subjects experiment was proposed for testing the performance of the NPST3 framework.

\begin{figure}[h]
  \vspace{1mm}
  \centering
  \includegraphics[width=\columnwidth]{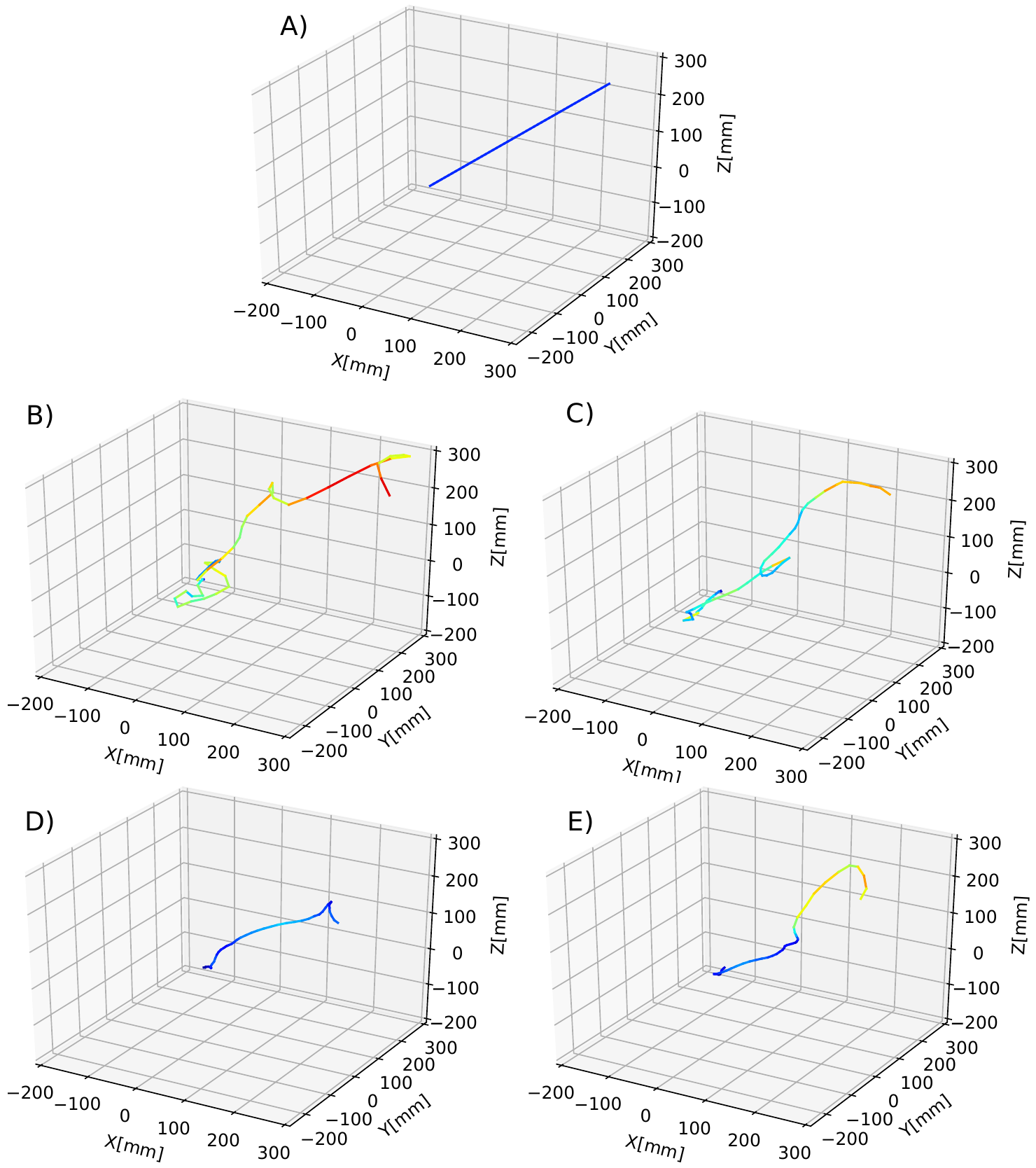} 
  \includegraphics[width=0.48\textwidth]{./figures/colormap.pdf}
    \vspace*{-6mm}
    \caption{Cartesian motions generated with the NPST3 algorithm. The trajectory at the top A) depicts the Content motion. The transferred Styles are: B) anger/annoyance, C) happiness/joy, D) calm/acceptance, and E) sadness/grief. The style trajectories are extracted from the right hand of a human demonstrator using a Vicon optical motion capture system. }
  \label{fig:trajectories}
  \vspace*{-3mm}
\end{figure}

\subsection{Subjects}

We enrolled 73 volunteers (43 males, 29 females, 1 prefer not to say; age from 18 to 76 years old). Subjects came from different parts of the world (10 nationalities, 4 countries of residence), they had different education backgrounds (from middle school diplomas to PhD) and working situation (active, unemployed, and retired).
Only few of them (12) had an experience in the robotics field. Subjects were able to carry out the experiment remotely.

\subsection{Methods and Task}

We considered a representative simple Content motion, a straight line, depicted at the top of Fig. \ref{fig:trajectories}. Then, we applied the four considered Styles following the methods described in Sec.~\ref{sec:framework}, yielding to four different motions.
For the sake of simplicity, the motions were generated offline, but the framework is capable of generating them at runtime, e.g., a user teleoperates a robotic manipulator and the chosen Style is applied immediately to the motion of the robot.
The four resulting motions were executed by a 7-DoF Franka Emika Panda robot, commanding velocities to the robot with respect to its end-effector frame.
Each motion was recorded through an external camera posed in front of the robot, resulting in four videos showing the same Content trajectory executed by the Franka robot in the four different Styles.
The video of these four executions is available as supplemental material and at \url{https://youtu.be/ynfx2hhfoUs}.

Subjects were presented with the four videos, with no information regarding the considered Styles nor about how the proposed framework works.
For each video, subjects were asked to write down, using one word, what kind of emotion the robot motion elicits in them.
Once their choices were registered, they were asked to watch the videos again. This time, subjects were asked to match each video with one of the considered Styles, i.e., anger/annoyance, happiness/joy, calm/acceptance, and sadness/grief, using a dropdown menu next to each video.
Subjects could watch the videos how many times they wanted.

\begin{figure}[h]
	\vspace{1mm}
	\centering
	\includegraphics[width=0.39\textwidth]{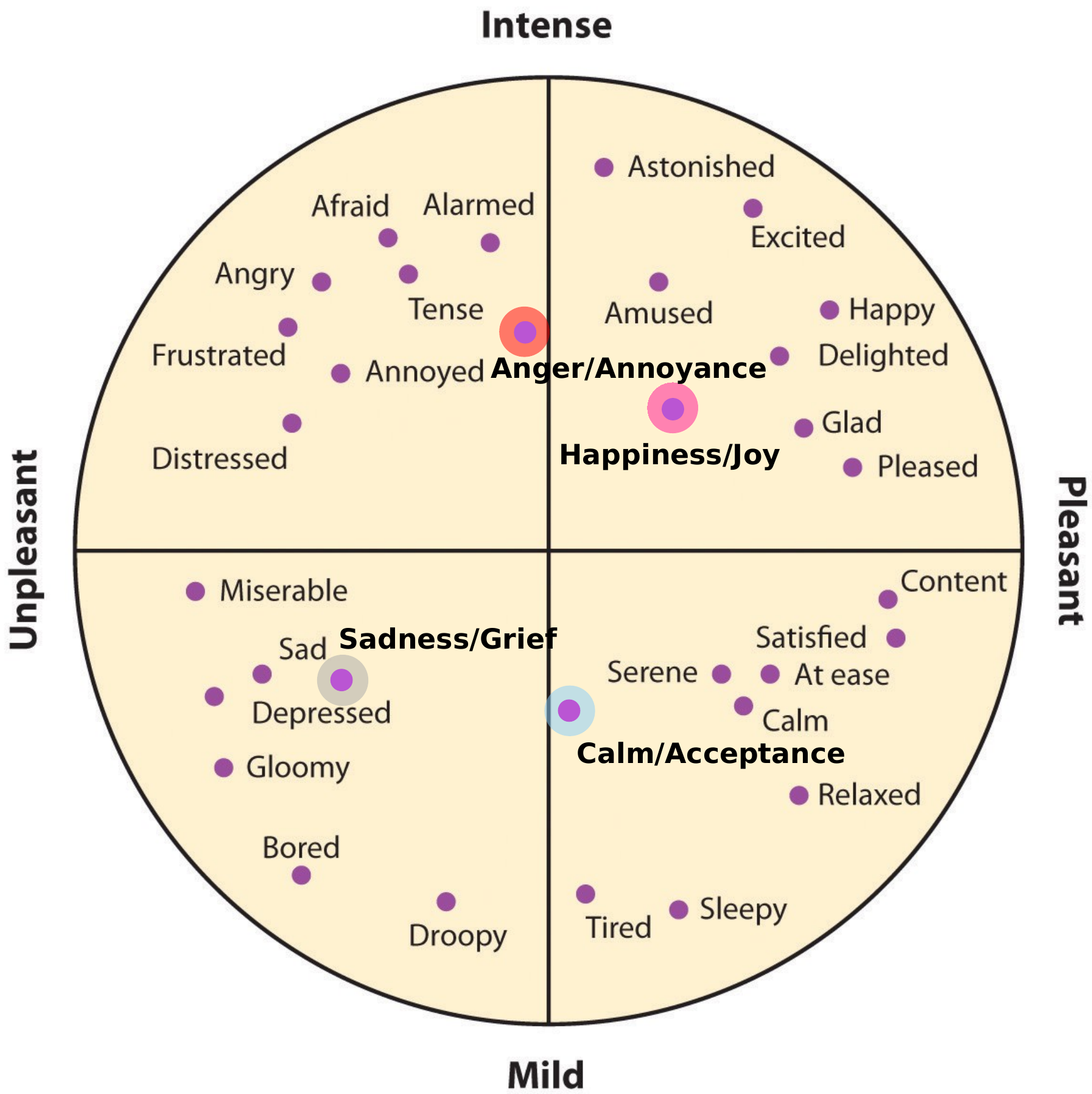} 
	\caption{Wheel of emotion. Considering the center of the wheel as the origin of a standard Cartesian coordinate system growing towards ``Intense'' (top) and ``Pleasant'' (right), we can give to each emotion in the wheel a coordinate, e.g., ``pleased'' can be at (14, 3) and ``tired'' at (-15, 1). Doing so, we can calculate the average answers of the subjects and place them in the wheel (bold). Figure inspired from Foxcroft \emph{et al.} \cite{Foxcroft2014} and based on the model proposed by Rusell \emph{et al.}~\cite{Rusell1980}.
	}
	\vspace*{-4mm}
	\label{fig:circumplex-model}
\end{figure}

\begin{figure}[h]
	\centering
	\includegraphics[width=0.47\textwidth]{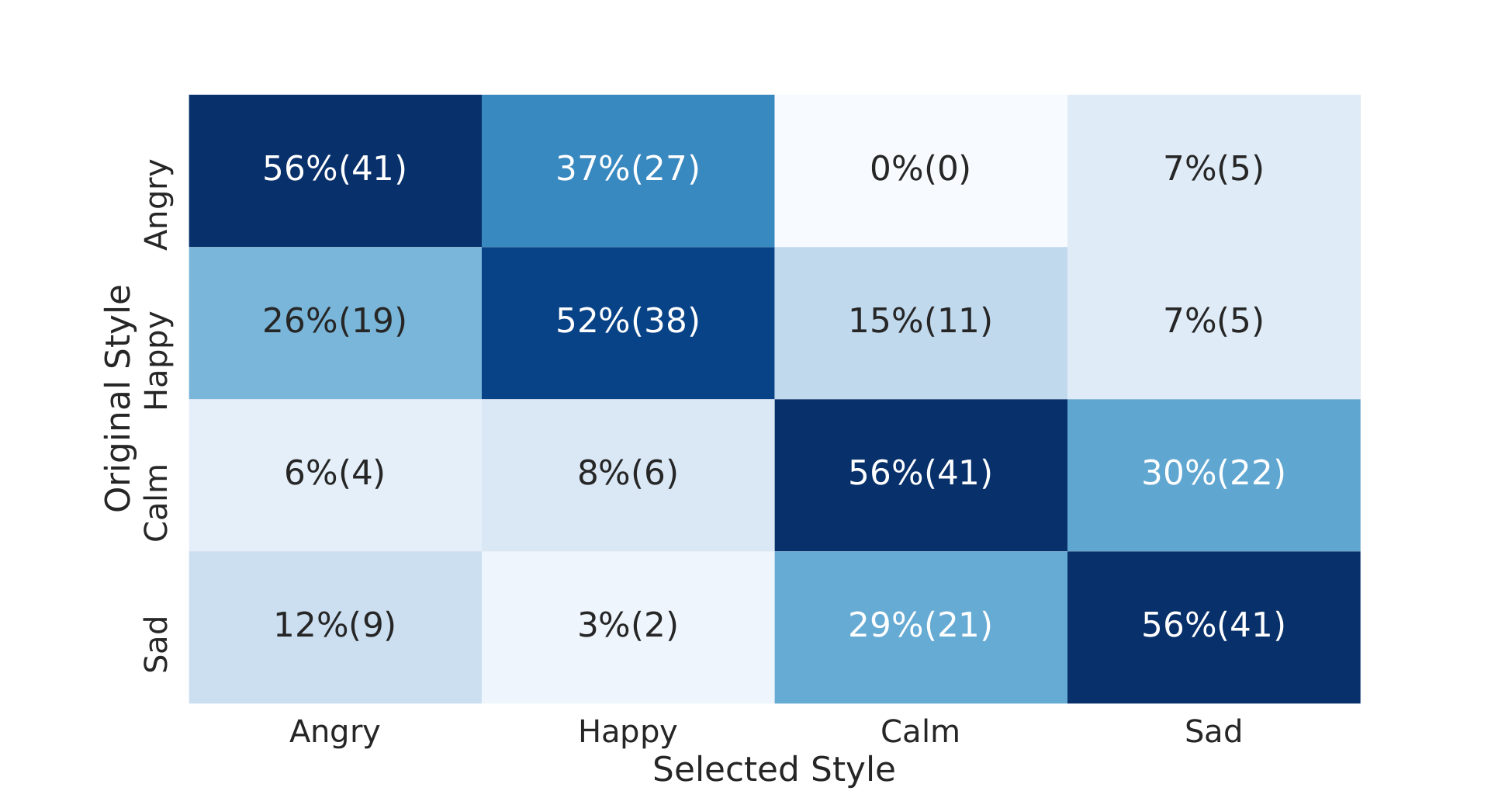} 
	\caption{Constrained multiple choice response. Each row depicts a different Style and associated video shown to the participants. Each column depicts the selected Style. The cells contain the percentage and number of times each Style was selected. All the videos were presented to all the volunteers. }
	\label{fig:results}
	\vspace*{-4mm}
\end{figure}

\begin{figure*}[h]
  \vspace*{2mm}
  \centering
 \hspace*{-2mm}    \includegraphics[width=0.48\textwidth]{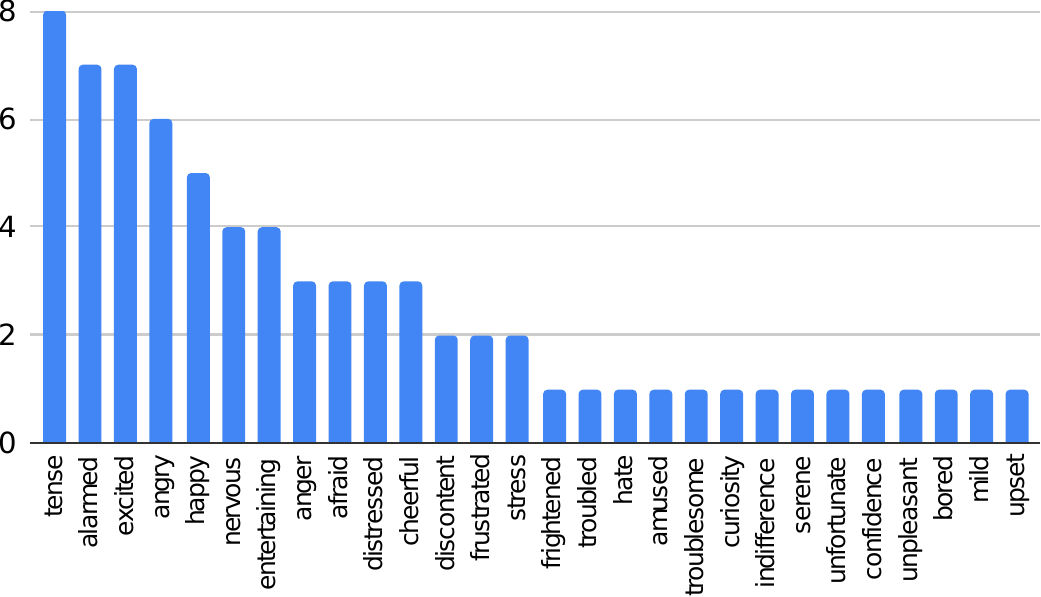}
 \includegraphics[width=0.48\textwidth]{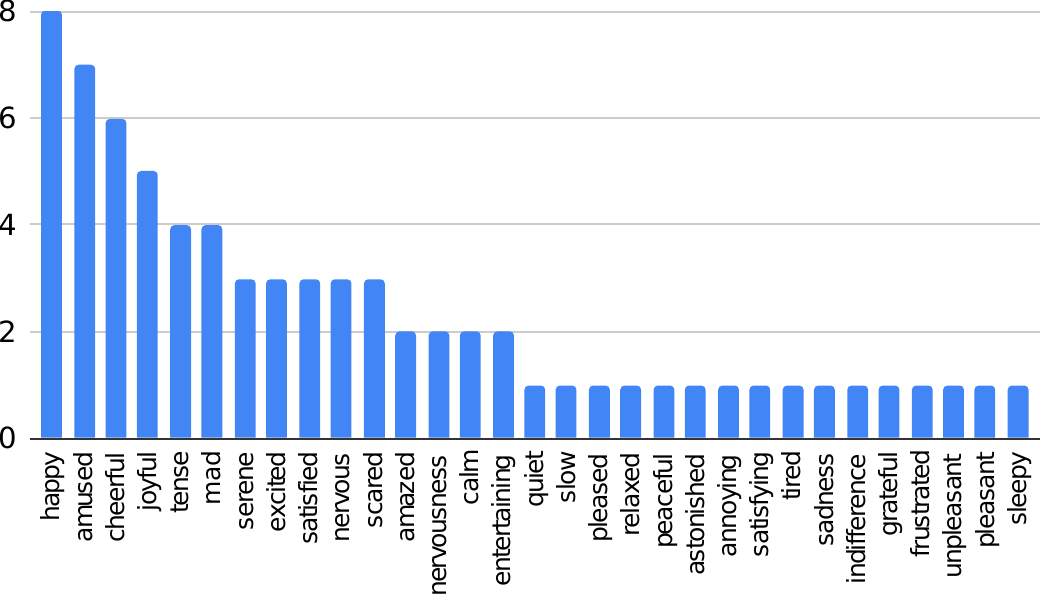}
    
    \vspace*{-6mm}
      \centering
\null\hfill   \subfloat[][Anger/annoyance]{     \includegraphics[width=0.27\textwidth]{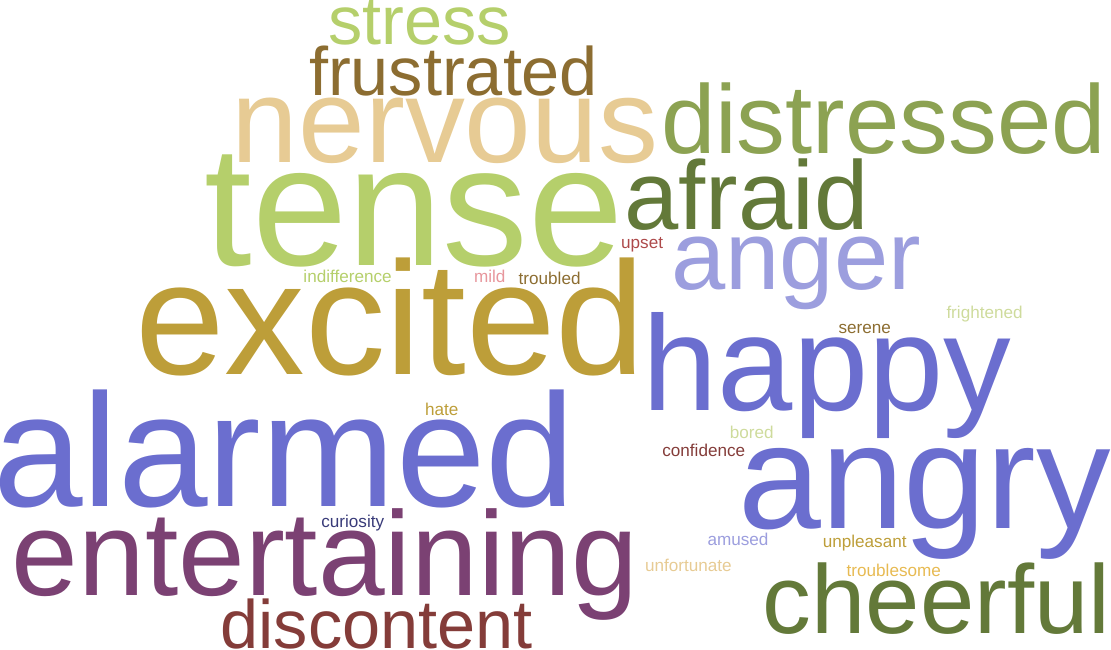}\label{angry}}
\hfill
  \subfloat[][Happiness/joy]{     \includegraphics[width=0.27\textwidth]{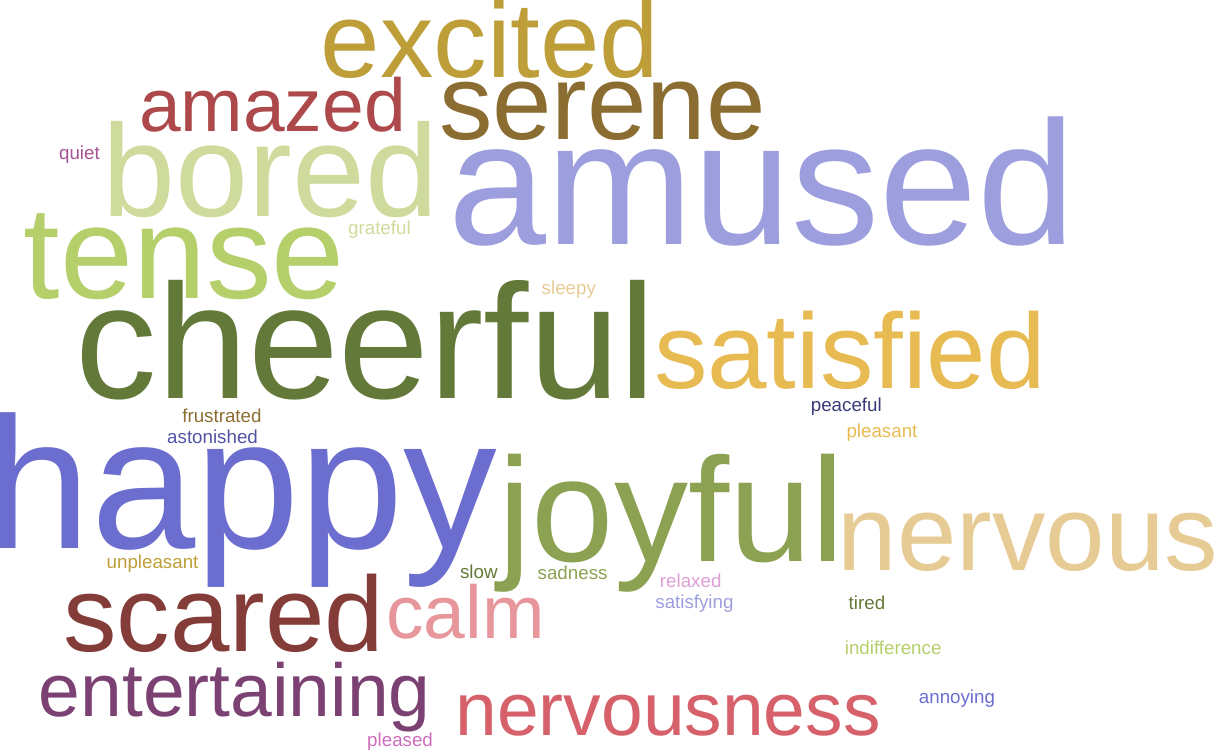}\label{happy}}
  \hfill\null
    
    \vspace*{8mm}
  \hspace*{-2mm}      \includegraphics[width=0.48\textwidth]{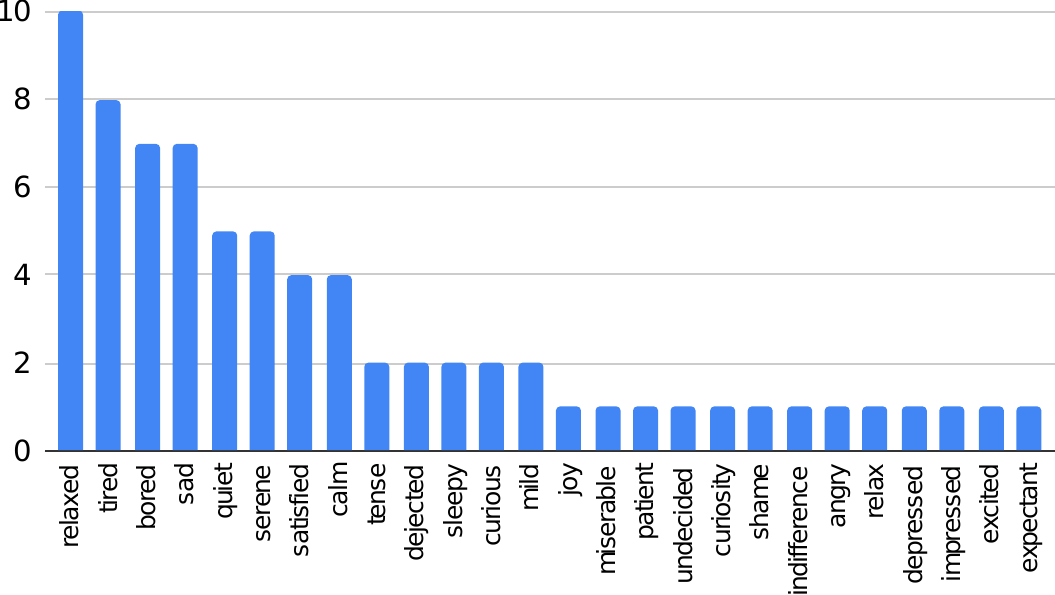}
    \includegraphics[width=0.48\textwidth]{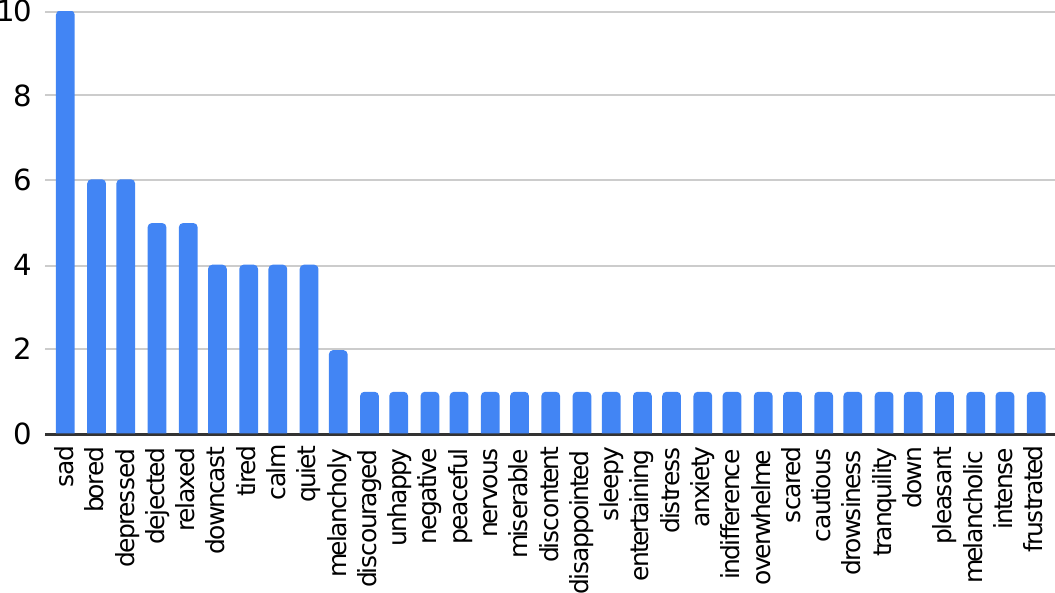}
    
  \vspace*{-5mm}
\null\hfill     \subfloat[][Calm/acceptance]{       \includegraphics[width=0.27\textwidth]{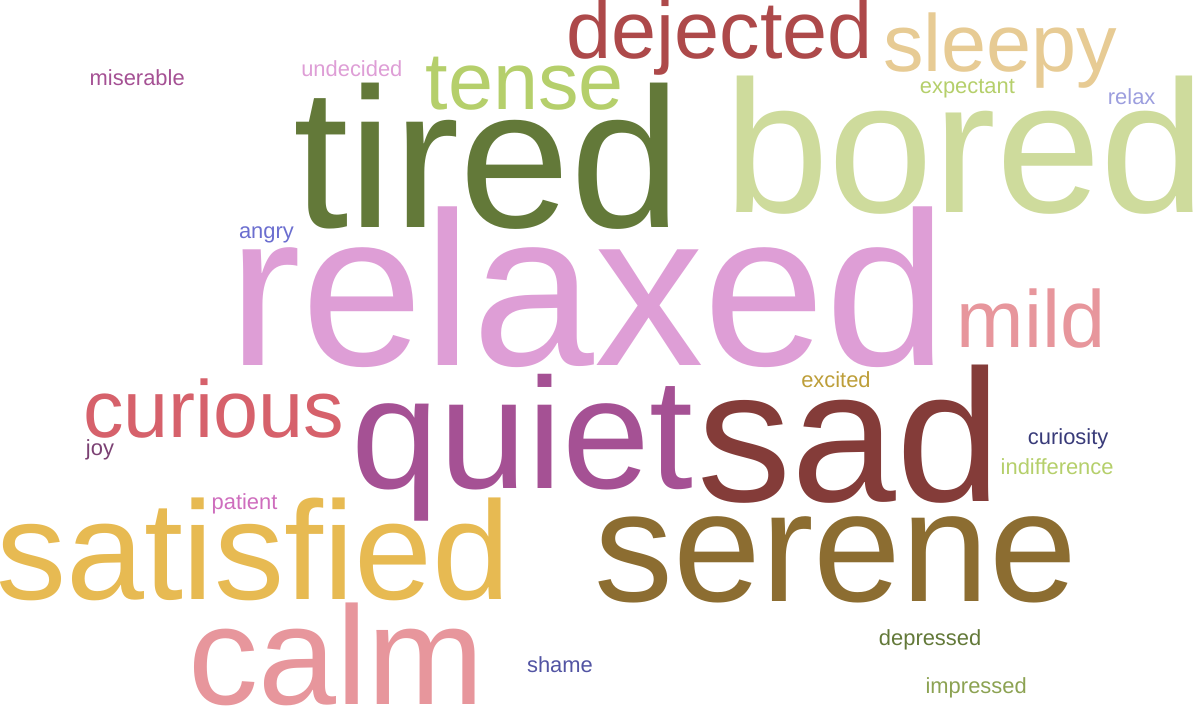}\label{calm}}
\hfill
 \subfloat[][Sadness/grief]{       \includegraphics[width=0.27\textwidth]{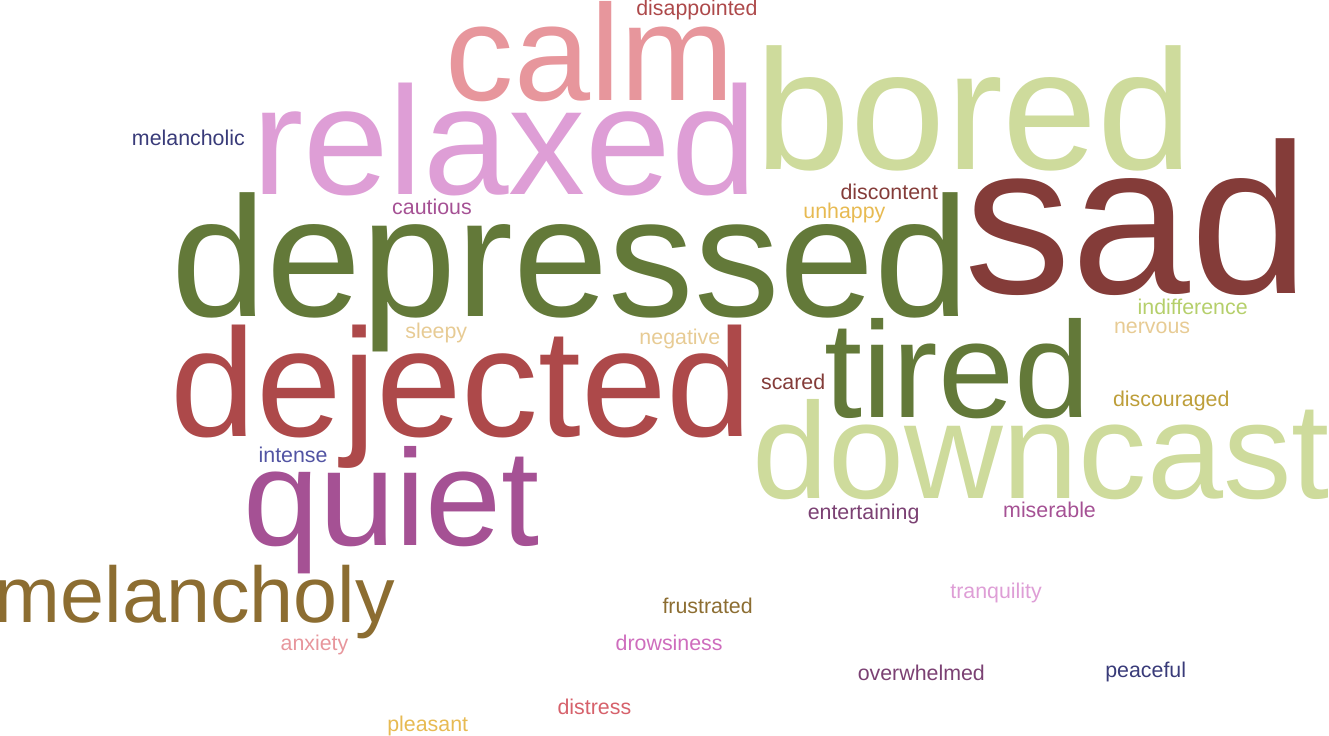}\label{sad}}
   \hfill\null

  \caption{Free text response. Each set of figures represents the results obtained with each of the four videos: \protect\subref{angry} anger/annoyance, \protect\subref{happy} happiness/joy, \protect\subref{calm} calm/acceptance, \protect\subref{sad} sadness/grief. The histograms show the frequency of the responses given by the subjects. For easier visualization, the same information is also reported using a word cloud (words in a larger font were more often chosen). The four motions were showed to all of the volunteers.}
  \vspace*{-3mm}
  \label{fig:free-results}
  
\end{figure*}

\subsection{Results}

Results are summarized in Figs. \ref{fig:circumplex-model}, \ref{fig:results} and  \ref{fig:free-results}.
Figs. ~\ref{fig:circumplex-model} and \ref{fig:free-results} shows the results of the first part of the questionnaire (free text response).
In Fig.~\ref{fig:circumplex-model}, the responses are placed within a wheel of emotions~\cite{Plutchik1980}. Each answered emotion is given a coordinate and the average response is reported in bold. We can see that users answered with emotions very close to the target one. In Fig.~\ref{fig:free-results}, emotions are ordered as a function of the frequency in which they appear. Finally, Fig.~\ref{fig:results} shows the results of the second part of the questionnaire (constrained multiple choice response).

\section{Discussion and Conclusions}

In this paper, NPST3 is proposed as a way to introduce Style Transfer in continuos robotic action spaces using DRL. 
The resulting network is designed to work with incomplete motion trajectories. This allows the online teleoperation of the robot during the Style Transfer process.

The results obtained with the experiments show how the proposed NPST3 framework is able to successfully transfer the original Style to the generated robotic motion.
Results in Figs.~\ref{fig:circumplex-model}, ~\ref{fig:results} and~\ref{fig:free-results} show that subjects were quite effective in identifying the correct emotion, even in the first free-text response part. 
In the second part, the main confusion was between anger vs. happiness and calm vs. sadness. While most subjects identified correctly the emotion (always~$>$50\%), those who did not confused happy/angry and calm/sad.

While there is still work to do in extending this approach to other types of robots (e.g., humanoids, mobile robots) as well as to other robotic motions (e.g., joint-by-joint Stylized motion), this work is, in our opinion, a promising starting point for introducing NST-based control in robotics.
In the next future, we plan to implement the proposed approach on humanoids, studying how the motion can be modified using Style Transfer as well as how such motion can be related to other features of the robot, e.g., its facial expression.

\section{Acknowledgment}
The research leading to these results has received funding from: RoboCity2030-DIH-CM, Madrid Robotics Digital Innovation Hub, S2018/NMT-4331, funded by “Programas de Actividades I+D en la Comunidad de Madrid” and cofunded by Structural Funds of the EU; ROBOASSET, "Sistemas rob\'oticos inteligentes de diagn\'ostico y rehabilitaci\'on de terapias de miembro superior", PID2020-113508RB-I00 funded by AGENCIA ESTATAL DE INVESTIGACION (AEI); and ``Programa propio de investigaci\'on convocatoria de movilidad 2020" from Universidad Carlos III de Madrid. 

\bibliographystyle{IEEEtran}
\bibliography{references.bib}

\end{document}